# Embedding Reliability Verification Constraints into Generation Expansion Planning

Peng Liu
School of Electrical Engineering
and Automation
Harbin Institute of Technology
Harbin, China
liupengnetl@gmail.com
liupengpwrs@hit.edu.cn

Lian Cheng
School of Electrical Engineering
and Automation
Harbin Institute of Technology
Harbin, China
22s106107@stu.hit.edu.cn

Benjamin P. Omell
The Institute for Design of
Advanced Energy Systems
National Energy Technology
Laboratory
Pittsburgh, United States
Benjamin.Omell@netl.doe.gov

Anthony P. Burgard
The Institute for Design of
Advanced Energy Systems
National Energy Technology
Laboratory
Pittsburgh, United States
Anthony.Burgard@netl.doe.gov

*Abstract*—Generation planning approaches face challenges in managing the incompatible mathematical structures between stochastic production simulations for reliability assessment and optimization models for generation planning, which hinders the integration of reliability constraints. This study proposes an approach to embedding reliability verification constraints into generation expansion planning by leveraging a weighted oblique decision tree (WODT) technique. For each planning year, a generation mix dataset, labeled with reliability assessment simulations, is generated. An WODT model is trained using this dataset. Reliability-feasible regions are extracted via depth-first search technique and formulated as disjunctive constraints. These constraints are then transformed into mixed-integer linear form using a convex hull modeling technique and embedded into a unit commitment-integrated generation expansion planning model. The proposed approach is validated through a long-term generation planning case study for the Electric Reliability Council of Texas (ERCOT) region, demonstrating its effectiveness in achieving reliable and optimal planning solutions.

## I. Introduction

In the transition to carbon-neutral power systems, the installed capacity of wind and solar energy is expected to grow significantly each year [1]. Recent operational insights reveal that in regions with high renewable energy penetration, maintaining reliable power supply is becoming increasingly challenging [2]. Historically, capacity inadequacies were localized and limited to specific periods. With increased renewable energy penetration, these challenges are expanding, with a tendency to affect the entire system across multiple operational periods [3]. Moreover, frequent forced outages of conventional and renewable generators, combined with renewable generation intermittency and extreme weather conditions—such as high temperatures with little or no wind (often coinciding with high evening peak demand and very low solar output) and overcast periods with reduced solar availability—further compound the challenges of maintaining supply-demand balance [4].

Generation Expansion Planning (GEP) is vital for supporting the transition of power systems towards carbon neutrality. Given the inherent variability of wind and solar power generation, GEP is often formulated as a unit commitment (UC)-integrated model [5], [6]. In our previous work [7], a mixed-integer linear programming model and a corresponding algorithm were developed to address the UC-integrated generation and transmission expansion planning problem. The GEP component of this model, derived from [6], incorporates reserve margin metrics to ensure both capacity adequacy and generation reliability. By integrating a reserve margin into the capacity constraints, the models in [6] and [7] ensures that the installed capacity of conventional generation and the credible capacity of renewable generation exceed the system's peak load by a specific margin annually [8], [9].

From the perspective of ensuring resource adequacy, the reserve margin-based approach, through simple, suffers from the following limitations: 1) The capacity constraints, based on the reserve margin, fail to incorporate forced outage rates and detailed time-series data spanning 8,760 hours or more [10], which are essential for capturing the intermittency of wind and solar generation, as well as scenarios involving severely constrained renewable generation with coincident peak loads; 2) Therefore, the reserve margin is typically set based on rules of thumb, often leading to redundancy and inefficiencies in generation planning. In [11], the authors developed a trial-and-error-based iterative approach for dynamic adjustment of annual planning reserve margin. While this method leverages forced outage rates and detailed time-series data to some extent, the reserve margin settings are still determined through incremental steps governed by empirical rules. As a result, it does not fully address the issues of capacity redundancy and inefficiencies inherent in the reserve margin-based approach.

This paper aims to leverage forced outage rates, detailed 8,760-hour (or more) time-series data, and generation reliability requirements (e.g., loss of load hour ≤ a specified threshold) to develop reliability verification constraints that replace traditional reserve margin-based capacity constraints in GEP models. The proposed approach resolves the inherent incompatibilities between two mathematical frameworks: (a) reliability assessments that employ time-resolved (day-by-day/hourly) simulations, and (b) GEP models that rely on large-scale optimization formulations. By solving the enhanced GEP model, the proposed approach seeks to formulate more efficient investment and retirement strategies that ensure system reliability, reduce redundancy, and optimize overall costs, thereby providing essential support for the secure transition of power systems to carbon neutrality.

The contributions of this paper are summarized as follows:
(1) Reliability verification constraints for each planning year are formulated using a weighted oblique decision tree (WODT) modeling technique, ensuring compatibility with the optimization framework of GEP models.
(2) The dimensionality of the feature variables and the training dataset for WODT is reduced by excluding generator types whose planning decisions are driven primarily by economic factors rather than reliability considerations, thereby ensuring computational feasibility.

This work was jointly supported by the NETL Research Program, funded by the U.S. Department of Energy and administered by the Oak Ridge Institute for Science and Education (grant number P-1-02822), and the Fundamental Research Funds for the Central Universities of China (grant number 5710011622). This paper has been accepted for presentation at the IEEE PES General Meeting 2025. © IEEE.

(3) A 20-year generation planning case in the Electric Reliability Council of Texas (ERCOT) region is studied to verify the effectiveness of the enhanced GEP model, which incorporates reliability verification constraints, in achieving reliable and optimal planning solutions.

The rest of the paper is summarized as follows. Section II describes construction of training dataset. Section III demonstrates the formulation and incorporation of reliability verification constraints. Section IV presents the case studies and discussions. Section V concludes the paper.

## II. TRAINING DATA CONSTRUCTION

Reliability verification constraints are formulated using a WODT modeling technique on an annual basis throughout the planning horizon. The WODT is modeled for each year using the potential operational quantities of certain types of generators as its feature variables. Training the WODT requires comprehensive datasets representing the potential operational quantities of various generation types. The term "potential operational quantity" is used to highlight that not all generators are available for operation due to unexpected forced outages.

### A. Feature Variable Selection

The GEP model usually includes various new and existing generation types (or technologies). However, using potential operational quantities of all generation types as feature variables results in high-dimensional WODTs, challenging data generation and training. To maintain computational feasibility, the potential operational quantities for new generation types with high costs and existing generation types with low life-extension costs are excluded from the feature variables, as their decisions are primarily influenced by economic factors rather than reliability. This approach reduces dimensionality while preserving accuracy of WODT.

Planning decisions for the aforementioned generation types are mainly driven by economic factors, resulting in little variation in their potential operational quantities under different planning reserve margin conditions. To identify these generation types, the GEP model can be configured with varying planning reserve margins. Solving the model then determines the generation types with unchanged potential operational quantities.

To this end, set $K$ distinct initial step sizes $\Delta_1^{res} \cdots \Delta_k^{res} \cdots \Delta_K^{res}$ with $\Delta_1^{res} < \cdots < \Delta_k^{res} < \cdots < \Delta_K^{res}$ for the adaptive increment of planning reserve margins. The potential operational quantities under different initial step sizes can be determined by using the process outlined in [11]. For each planning year, these decision values for new and existing generation types are compiled into two matrices, as shown in (1)-(2).

$$\begin{bmatrix} ngo^{new\_op}_{\Delta_1^{res},\overline{[1]},t} & \cdots & ngo^{new\_op}_{\Delta_k^{res},\overline{[1]},t} & \cdots & ngo^{new\_op}_{\Delta_K^{res},\overline{[1]},t} \\ \vdots & \ddots & \vdots & \ddots & \vdots \\ ngo^{new\_op}_{\Delta_1^{res},\overline{[i]},t} & \cdots & ngo^{new\_op}_{\Delta_k^{res},\overline{[i]},t} & \cdots & ngo^{new\_op}_{\Delta_K^{res},\overline{[i]},t} \\ \vdots & \ddots & \vdots & \ddots & \vdots \\ ngo^{new\_op}_{\Delta_1^{res},\overline{[\mathcal{T}^{new}]},t} & \cdots & ngo^{new\_op}_{\Delta_k^{res},\overline{[\mathcal{T}^{new}]},t} & \cdots & ngo^{new\_op}_{\Delta_K^{res},\overline{[\mathcal{T}^{new}]},t} \end{bmatrix} \quad (1)$$

$$\begin{bmatrix} ngo^{old\_op}_{\Delta_1^{res},\overline{[1]},t} & \cdots & ngo^{old\_op}_{\Delta_k^{res},\overline{[1]},t} & \cdots & ngo^{old\_op}_{\Delta_K^{res},\overline{[1]},t} \\ \vdots & \ddots & \vdots & \ddots & \vdots \\ ngo^{old\_op}_{\Delta_1^{res},\overline{[i]},t} & \cdots & ngo^{old\_op}_{\Delta_k^{res},\overline{[i]},t} & \cdots & ngo^{old\_op}_{\Delta_K^{res},\overline{[i]},t} \\ \vdots & \ddots & \vdots & \ddots & \vdots \\ ngo^{old\_op}_{\Delta_1^{res},\overline{[\mathcal{T}^{old}]},t} & \cdots & ngo^{old\_op}_{\Delta_k^{res},\overline{[\mathcal{T}^{old}]},t} & \cdots & ngo^{old\_op}_{\Delta_K^{res},\overline{[\mathcal{T}^{old}]},t} \end{bmatrix} \quad (2)$$

In (1) and (2), the rows correspond to different generation types, while the columns represent the respective initial step sizes used for the adaptive increment of planning reserve margins; $t \in \{1,\cdots,T\}$, $T$ is the number of years within the planning horizon; $ngo^{new\_op}_{\Delta_k^{res},\overline{[i]},t}$ is the decision value of potential operational quantity for the $i^{th}$ type of new generators; $\mathcal{T}^{new}$ is the set including names of all types of new generators; $\overline{[i]}$ is the name of the $i^{th}$ type of new generators, $\overline{[i]} \in \mathcal{T}^{new}$. Accordingly, $ngo^{old\_op}_{\Delta_k^{res},\overline{[i]},t}$ is the decision value of potential operational quantity for the $i^{th}$ type of existing generators; $\mathcal{T}^{old}$ is the set including names of all types of existing generators; $\overline{[i]}$ is the name of the $i^{th}$ type of old generators, $\overline{[i]} \in \mathcal{T}^{old}$.

For each planning year, each row of (1) and (2) is analyzed to identify new generation types with consistently zero potential operational quantities and existing generation types with constant values equal or close to their initial quantities. These generation types are classified into the sets $\mathcal{T}^{nf\_new}$ and $\mathcal{T}^{nf\_old}$. Since the generation types in these sets consistently remain outside the investment portfolio or are not retired under varying planning reserve margins influenced by different initial step sizes, it can be inferred that their planning decisions are primarily determined by cost factors. These sets are excluded from $\mathcal{T}^{new}$ and $\mathcal{T}^{old}$, resulting in refined sets $\mathcal{T}^{f\_new}$ and $\mathcal{T}^{f\_old}$. Potential operational quantities of generator types in $\mathcal{T}^{f\_new}$ and $\mathcal{T}^{f\_old}$ serve as the feature variables of WODT for each year.

### B. Feature Variable Bound Setup

For each planning year $t$ ($t \in \{1,\cdots,T\}$), it is essential to set appropriate upper and lower bounds for feature variables. The bounds are introduced for two purposes: to manage training sample size and to ensure the convex hull model accurately represents the disjunction of trained WODT — a requirement for theoretical accuracy in mixed-integer linear formulations (see Section III-A for further analysis). Determining these bounds requires balancing the generalization capability of WODT with computational tractability, allowing for a certain degree of flexibility. However, the bounds are not entirely arbitrary and should align with basic guidelines.

For ease of description, the lower and upper bounds of potential operation quantity for the $i^{th}$ type of new generator in $\mathcal{T}^{f\_new}$ are denoted as $[ngo^{f\_new}_{\overline{[i]},t,lower}, ngo^{f\_new}_{\overline{[i]},t,upper}]$, and those for the $i^{th}$ type of existing generator in $\mathcal{T}^{f\_old}$ are denoted as $[ngo^{f\_old}_{\overline{[i]},t,lower}, ngo^{f\_old}_{\overline{[i]},t,upper}]$. $\overline{[i]}$ is the name of the $i^{th}$ type of new generator in $\mathcal{T}^{f\_new}$. $\overline{[i]}$ is the name of the $i^{th}$ type of existing generator types in $\mathcal{T}^{f\_old}$.

For each planning year $t$ ($t \in \{1,\cdots,T\}$), to ensure the training datasets comprehensively capture the variability in generation reliability metrics (LOLH) across diverse generation mix, it is recommended to setup the upper and lower bounds for the potential operational quantities for new and existing generation types in $\mathcal{T}^{f\_new}$ and $\mathcal{T}^{f\_old}$ in accordance with the constraints specified by (3) to (6):

$$ngo^{f\_new}_{\overline{[i]},t,lower} \leq ngo^{f\_new}_{\overline{[i]},t,min} \quad (3)$$

$$ngo^{f\_new}_{\overline{[i]},t,upper} \geq ngo^{f\_new}_{\overline{[i]},t,max} \quad (4)$$

$$ngo^{f\_old}_{\overline{[i]},t,lower} \leq ngo^{f\_old}_{\overline{[i]},t,min} \quad (5)$$

$$ngo^{f\_old}_{\overline{[i]},t,upper} \geq ngo^{f\_old}_{\overline{[i]},t,max} \quad (6)$$

In (3)-(6), $ngo^{\text{f\_new}}_{\overline{[i]},t,\min}$ and $ngo^{\text{f\_new}}_{\overline{[i]},t,\max}$ represent the minimum and maximum values of the $i^{\text{th}}$ type of new generator in $\mathcal{T}^{\text{f\_new}}$, as determined by the process described in [11] with reserve margins gradually increased from different initial step sizes $\Delta^{\text{res}}_1 \cdots \Delta^{\text{res}}_k \cdots \Delta^{\text{res}}_K$. $ngo^{\text{f\_old}}_{\overline{[i]},t,\min}$ and $ngo^{\text{f\_old}}_{\overline{[i]},t,\max}$ represent the minimum and maximum values of the $i^{\text{th}}$ type of existing generator in $\mathcal{T}^{\text{f\_old}}$, which are obtained under these distinct initial step sizes $\Delta^{\text{res}}_1 \cdots \Delta^{\text{res}}_k \cdots \Delta^{\text{res}}_K$.

### C. Training Dataset Construction

Data samples are generated by iteratively enumerating feature variable values within their bounds. These values, along with potential operational quantities of generation types in $\mathcal{T}^{\text{nf\_new}}$ and $\mathcal{T}^{\text{nf\_old}}$ (specified in Section II-A), forced outage rates, and historical time-series data, are input into the reliability assessment simulation procedure described in [11]. This procedure evaluates loss of load hour (LOLH) metrics to assess compliance with NERC's "1/10" reliability criterion (i.e., LOLH $\leqslant$ 2.4). Samples that meet this criterion are labeled with '1', while others are labeled with '0'.

## III. RELIABILITY VERIFICATION CONSTRAINTS FORMULATION AND INCORPORATION

### A. Reliability Verification Constraints Formulation

In each planning year, the WODT is trained using labeled dataset through a recursive strategy based on the Limited-memory Broyden–Fletcher–Goldfarb–Shanno (L-BFGS) algorithm, as detailed in [12]. The depth-first search explores all leaf nodes marked '1', constructing linear disjunctive constraints (logical "or" operations, as described in [13]) that delineate the reliability-feasible regions as follows:

$$\bigvee_{k \in \{1,\cdots,N^{\text{fes}}_t\}} \begin{bmatrix} W_{t,k} \\ \boldsymbol{R}_{t,k}\, \boldsymbol{x}^{\text{f}}_t \leq \boldsymbol{b}_{t,k} \\ \boldsymbol{x}^{\text{f}}_{t,\text{lower}} \leq \boldsymbol{x}^{\text{f}}_t \leq \boldsymbol{x}^{\text{f}}_{t,\text{upper}} \end{bmatrix} \quad \forall t \in \{1,\cdots,T\} \quad (7)$$

where $\vee$ denotes logical "or" operator, $W_{t,k}$ is a binary variable, representing the selection of the $k^{\text{th}}$ reliability-feasible region or not. $\boldsymbol{R}_{t,k}$ and $\boldsymbol{b}_{t,k}$ represent the coefficient matrix and constant vector for the $k^{\text{th}}$ reliability-feasible region. $N^{\text{fes}}_t$ is the number of reliability-feasible regions in the $t^{\text{th}}$ planning year. $\boldsymbol{x}^{\text{f}}_t$ is the vector of feature variables. $\boldsymbol{x}^{\text{f}}_{t,\text{lower}}$ and $\boldsymbol{x}^{\text{f}}_{t,\text{upper}}$ are vectors formulated by the lower and upper bounds of each feature variable. The disjunctive constraints of (7) can be reformulated into mixed-integer linear constraints using the convex hull modeling technique [14], as shown in (8) to (11):

$$\sum_{k \in \{1,\cdots,N^{\text{fes}}_t\}} \boldsymbol{z}^{\text{f}}_{t,k} = \boldsymbol{x}^{\text{f}}_t \quad (8)$$

$$\sum_{k \in \{1,\cdots,N^{\text{fes}}_t\}} W_{t,k} = 1 \quad (9)$$

$$\boldsymbol{R}_{t,k}\, \boldsymbol{z}^{\text{f}}_{t,k} \leq \boldsymbol{b}_{t,k} W_{t,k} \quad (10)$$

$$W_{t,k}\, \boldsymbol{x}^{\text{f}}_{t,\text{lower}} \leq \boldsymbol{z}^{\text{f}}_{t,k} \leq W_{t,k}\, \boldsymbol{x}^{\text{f}}_{t,\text{upper}} \quad (11)$$

where $\boldsymbol{z}^{\text{f}}_{t,k}$ denotes the auxiliary variable, $k \in \{1,\cdots,N^{\text{fes}}_t\}$, $t \in \{1,\cdots,T\}$. These constraints ensure that the generation mix for each planning year meets the reliability criteria. The recession cone associated with each disjunct in (7) is reduced to the origin. Thus, (8)-(11) can exactly represent (7) without introducing approximation errors, as detailed in Section 2.1.2 of [14].

### B. Reliability Verification Constraint Integration

The constraints (8)-(11) can be integrated into the GEP model described in [6] and [7]. It is very important to note that the labels for data samples are determined under the condition that the potential operational quantities of generators in $\mathcal{T}^{\text{nf\_new}}$ and $\mathcal{T}^{\text{nf\_old}}$ are fixed at the values specified in Section II-A. To ensure the GEP model work correctly with the embedded reliability verification constraints, decision variables for these quantities should remain fixed. Additionally, because the GEP model already incorporates reliability verification constraints, there is no need to setup planning reserve margins. The constraints (8)-(11) involve only investment decision variables, which do not disrupt the structure of the original GEP model. The enhanced model can be solved using the Benders decomposition technique described in [6] and [15].

## IV. CASE STUDIES AND DISCUSSION

This proposed approach is validated using a case study of 20-year generation planning in the region of Electric Reliability Council of Texas (ERCOT).

### A. Required Data Description

Detailed generation parameters are described in [6] and [7]. The data on forced outage rates are presented in Table I, which are sourced from [16] and [17]. The generation types corresponding to the terminologies in Table I are as follows: **PV**: photovoltaic; **wind**: wind turbine; **coal-st**: coal-fired steam turbine; **nuc-st**: nuclear-fired steam turbine; **ng-{cc, ct, st, cc-ccs}**: natural gas-fired {combined cycle, combustion turbine, steam turbine, combined cycle with carbon capture}. The annual load growth rate is set to be 1.4%. Time-series data over 8,760 hours for wind and solar capacity factors, along with historical load demand, are utilized to evaluate the LOLH metrics. These data are obtained from [18] and [19]. The capacity factors, defined as the ratio of total generation to installed capacity, reflect the wind and solar output characteristics across the entire ERCOT region.

TABLE I
FORCED OUTAGE RATE

| Generation type | PV | wind | coal-st | nuc-st |
|---|---|---|---|---|
| FOR | 0.00055 | 0.017 | 0.079 | 0.035 |
| **Generation type** | **ng-cc** | **ng-ct** | **ng-st** | **ng-cc-ccs** |
| FOR | 0.0527 | 0.0311 | 0.079 | 0.0527 |

### B. Feature Variable Selection

The initial step sizes for adjusting reserve margins are set sequentially to 0.01, 0.02, 0.03, 0.04, and 0.05. The process described in [11] is then implemented. Five types of new generators—**wind-new**, **PV-new**, **ng-cc-new**, **ng-cc-ccs-new**, and **ng-ct-new**—are selected to formulate the set $\mathcal{T}^{\text{f\_new}}$, while two types of existing generators—**ng-cc-old** and **ng-st-old**—are chosen for the set $\mathcal{T}^{\text{f\_old}}$. The variables of potential operational quantities for these generation types are used as feature variables for the WODT in each planning year.

### C. Training Data Generation and Analysis

To generate the training dataset, the planning years requiring reliability verification constraints are first identified. This is achieved by examining the LOLH metrics associated with the planning decisions when the planning reserve margins are set to zero for each year. The results are listed in Table II.

From Table II, it can be observed that when the planning reserve margins are set to zero, the LOLH metrics for years 1 to 4 and years 18 to 20 are significantly unfavorable, substantially exceeding 2.4. In this study, reliability verification constraints are formulated for the first four and last three years of the planning horizon to evaluate their effectiveness in improving generation reliability.

TABLE II
LOLH FROM FOR EACH PLANNING YEAR (RESERVE MARGIN = 0)

| Year | 1 | 2 | 3 | 4 | 5 |
|---|---|---|---|---|---|
| LOLH | 44.58 | 69.53 | 51.80 | 101.45 | 0.16 |
| Year | 6 | 7 | 8 | 9 | 10 |
| LOLH | 0.05 | 0.10 | 0.20 | 0.46 | 4.84 |
| Year | 11 | 12 | 13 | 14 | 15 |
| LOLH | 12.38 | 19.11 | 22.56 | 23.06 | 25.62 |
| Year | 16 | 17 | 18 | 19 | 20 |
| LOLH | 32.37 | 67.67 | 143.31 | 194.58 | 290.83 |

The lower and upper bounds of each feature variable are set in compliance with (3)-(6). Under varying reserve margins driven by different initial steps, the minimum and maximum potential operational quantities of new generators in years 18 to 20 are significantly higher compared to years 1 to 4. To limit the sample size, conservative boundary settings are applied to feature variables for years 18 to 20. Conversely, the boundary settings for the first four years are more aggressive.

For example, for **wind-new**, the intervals between the minimum and maximum potential operational quantities under different initial step sizes for years 1 to 4 are [16, 16], [31, 31], [47, 47], and [78, 78], respectively. For years 18 to 20, the range is consistent at [432, 435]. The data generation boundary for **wind-new** in years 18 to 20 is set to [430, 444], with the lower bound reduced by 0.46% and the upper bound increased by 2%. For years 1 to 4, the data generation boundaries are set more aggressively at [0, 24], [16, 40], [32, 56], and [62, 86], with the lower bounds reduced by 100%, 48%, 32%, and 21%, respectively, and the upper bounds increased by 50%, 29%, 19%, and 10%, respectively.

In the first four-years, 1,455,300, 3,294,000, 4,026,000, and 6,222,000 samples were generated, respectively. In each of the last 3 years, 6,048,000 samples were generated. Figure 1(a) presents the boxplot of LOLH values for these samples, showing that the interquartile range (IQR) and the 25%-75% range are narrower in the last three years due to conservative boundary settings. This also results in lower maximum and mean LOLH values. Despite this, a large number of samples with both 0 and 1 labels are still observed in the final three years, as shown in Figure 1(b). This diversity significantly enhances the accuracy of the reliability verification constraints.

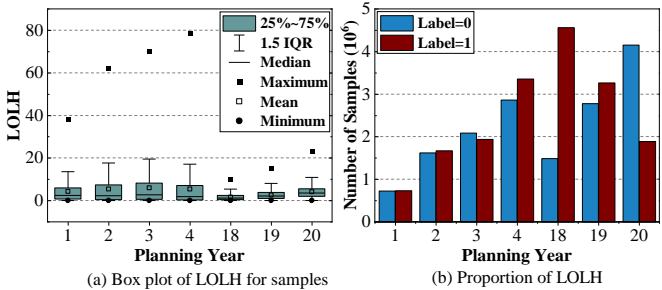

Fig.1 Comparison of actual capacity margins
(a) Box plot of LOLH for samples
(b) Proportion of LOLH

### D. Results and Discussion

The maximum depth of the WODT is set to 6, and the maximum number of iterations for the L-BFGS algorithm is set to 100. No theoretical method exists to guide the selection of these parameters. They are set via a trial-and-error process, ensuring that the ratio of samples assigned correctly to the leaf nodes by the trained WODT (samples labeled '1' are allocated to nodes labeled '1' and those labeled '0' are assigned to nodes labeled '0') exceeds 99.9%. Disjunctive constraints are extracted from each trained WODT. The number of polyhedral contained within each disjunction is 13, 20, 23, 25, 20, 19, and 19, respectively. Due to space limitations, detailed data are not provided. These disjunctions are then formulated into mixed-integer linear constraints in the form of (8)-(11) and embedded into the GEP model described in [6] and [7].

For clarity, the GEP method that incorporates reliability verification constraints is denoted as RVC-GEP. The method that employs incremental planning reserve margins, with an initial step size of 0.01, is referred to as RM-GEP. The LOLH values for each planning year under RVC-GEP are presented in Table III. A comparison between Tables II and III indicates that reliability verification constraints effectively reduce the LOLH metrics to meet the '1/10' standard. For example, in the first year, the LOLH decreases from 44.58 to 2.34, with similar improvements observed across the other planning years.

TABLE III
LOLH FROM FOR EACH PLANNING YEAR (RVC-GEP)

| Year | 1 | 2 | 3 | 4 | 5 |
|---|---|---|---|---|---|
| LOLH | 2.34 | 2.40 | 2.16 | 2.26 | 0.04 |
| Year | 6 | 7 | 8 | 9 | 10 |
| LOLH | 0 | 0 | 0 | 0 | 0 |
| Year | 11 | 12 | 13 | 14 | 15 |
| LOLH | 0 | 0.0014 | 0.011 | 0.028 | 0.077 |
| Year | 16 | 17 | 18 | 19 | 20 |
| LOLH | 0.15 | 0.36 | 0.94 | 1.96 | 2.40 |

Fig. 2 compares the actual capacity margins of the RVC-GEP and RM-GEP methods. As shown in Fig. 2(a), the RVC-GEP method achieves a substantial reduction in capacity margins compared to the RM-GEP method during the final four years of the planning horizon, with decreases of 0.0317, 0.0559, 0.0573, and 0.0479, respectively. Fig. 2(b) illustrates that the maximum, mean, median, first, and third percentile values of capacity margins for each planning year under the RVC-GEP method are consistently lower than those under the RM-GEP method. Fig. 2 and Table III demonstrate that the GEP model with reliability verification constraints effectively reduces redundant capacity while still meeting reliability requirements. The RVC-GEP method reduces total investment costs by $1.66 billion compared to the RM-GEP method by minimizing capacity redundancy. However, this reduction is partially offset by an increase in operational costs of $0.95 billion over the planning period, resulting in a net cost saving of $0.71 billion.

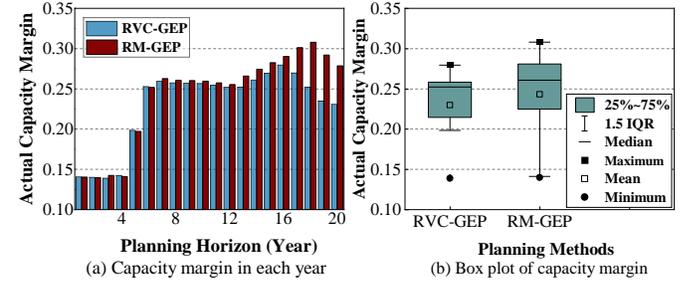

Fig.2 Comparison of actual capacity margins
(a) Capacity margin in each year
(b) Box plot of capacity margin

Fig. 3-(a) illustrates the installed capacity of generators by primary energy type for each planning year. Wind and solar generator capacities steadily increase over the planning period. During the first decade, gas-fired generation capacity rises gradually; however, due to increasing carbon taxes and the retirement of aging units, gas capacity begins to decline after the 10[th] year. Starting in the 16[th] year, it increases to meet growing demand. Coal-fired generation capacity remains stable in the first decade but gradually decreases in the following

decade due to retirements and higher carbon taxes. Throughout the planning period, nuclear generation capacity remains constant at 5164 MW.

Fig. 3-(b) presents the generation schedule for the day with the annual peak demand in the 20[th] year under *N*-3 contingency conditions. Three nuclear generators, with a combined capacity of 3.81 GW, are assumed to be offline. On this day, wind and solar generation are limited, each contributing only 12% of the total demand. The RVC-GEP method effectively develops a generation plan to ensure sufficient capacity under these extreme conditions. In this scenario, gas-fired generators serve as the primary power source, consistently maintaining high output levels and averaging 88% of their installed capacity throughout the day. Coal-fired generators begin operation in the 13[th] hour, generally maintaining low output levels except between the 19[th] and 23[rd] hours. On average, their output over the 24-hour period accounts for only 19% of their installed capacity. During the 21[st] hour, PV output drops to nearly zero, causing coal-fired generation to peak at 14.33 GW (91% of installed capacity), while gas-fired units reach their maximum output of 62.8 GW (100% of installed capacity). Due to limited ramp-up capability and high startup costs, nuclear-fired generators operate continuously at their full capacity of 1.35 GW throughout the day.

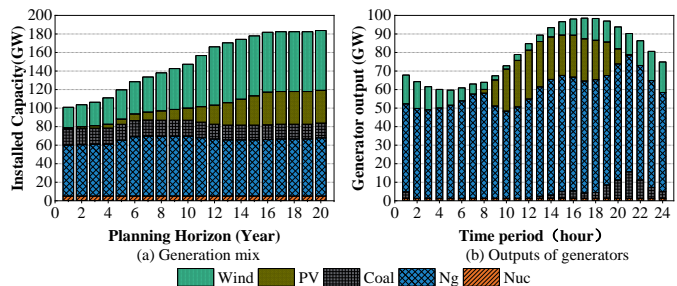

Fig. 3 Generation mix and outputs of generators under annual peak day

## V. Conclusion

This paper proposes an approach to incorporate reliability verification constraints into the GEP model using a WODT technique, addressing the incompatibility between reliability assessment simulations and GEP optimizations. Economy-dominated planning decisions are excluded from feature variables of WODT to reduce dimensionality. Reliability verification constraints enhance GEP models by capturing impacts of forced outages, renewable intermittency, and concurrent renewable inadequacy and peak loading conditions. Case studies demonstrate that the enhanced GEP model effectively optimizes planning decisions to meet reliability standards, minimize costs, and reduce capacity redundancy. The work presented in this paper supports the transition of power systems toward carbon-neutral energy paradigms.

The approach can be extended to account for uncertainties in long-term load forecasting. Future work will generate training data labels by using stochastic load-growth scenarios and integrate these scenarios with reliability verification constraints into the GEP framework. A more advanced decomposition algorithm will be developed to further enhance computational efficiency.